%% file: Template.tex
\documentclass{article}
\usepackage{spconf,amsmath,graphicx}

\usepackage{booktabs} 
\usepackage{tabularx}
\usepackage{enumitem}
\usepackage{url}
\usepackage{cite}
\usepackage{ifpdf}
\usepackage{algorithmic}
\usepackage{multirow}
\usepackage{subcaption}
\title{Gradient-Based Severity Labeling for Biomarker Classification in OCT}
\name{Kiran Kokilepersaud$^{\star}$ \enskip Mohit Prabhushankar$^{ \star}$ \enskip Ghassan AlRegib$^{\star}$ \enskip Stephanie Trejo Corona$^{\dagger}$ \enskip Charles Wykoff$^{\dagger}$ \enskip } 

\address {$^{\star}$ OLIVES at the Center for Signal and Information Processing CSIP,\\ 
School of Electrical and Computer Engineering, Georgia Institute of Technology, Atlanta, GA, USA \\
    $^{\dagger}$ Retina Consultants Texas, Retina Consultants of America, Houston, Texas, USA \\\{kpk6, mohit.p, alregib\}@gatech.edu             \{stephanie.trejo, ccwmd\}@retinaconsultantstexas.com}

\begin{document}
\twocolumn[{%

{ \large
\begin{itemize}[leftmargin=2.5cm, align=parleft, labelsep=3cm, itemsep=4ex,]

\item[\textbf{Citation}]{K. Kokilepersaud, M. Prabhushankar, G. AlRegib, S. Trejo Corona, C. Wykoff "Gradient-Based Severity Labeling for Biomarker Classification in OCT," in \textit{2022 IEEE International Conference on Image Processing (ICIP), Bordeaux, France, 2022.}}

\item[\textbf{Review}]{Date of Acceptance: June 6th 2022}

\item[\textbf{Codes}]{\url{https://github.com/olivesgatech/Severity_Contrastive_Learning.git}}

\item[\textbf{Bib}]  {@inproceedings\{kokilepersaud2022severity,\\
    title=\{Gradient-Based Severity Labeling for Biomarker Classification in OCT\},\\
    author=\{Kokilepersaud, Kiran and Prabhushankar, Mohit and AlRegib, Ghassan and Trejo Corona, Stephanie and Wykoff, Charles\},\\
    booktitle=\{IEEE International Conference on Image Processing\},\\
    year=\{2022\}\}}


\item[\textbf{Contact}]{
\{kpk6, mohit.p, alregib\}@gatech.edu\\\url{https://ghassanalregib.info/}\\}

\item[\textbf{Corresponding \\ Author}]{
alregib@gatech.edu}
\end{itemize}

}}]
\maketitle
\vspace{-1.8cm}
\begin{abstract}
\input{Second_Drafts/abstract}
\end{abstract}
\vspace{-.3cm}
\section{Introduction}
\label{sec:intro}

\input{Second_Drafts/introduction}


\section{RELATION TO PRIOR WORK}
\label{sec:prior}

\input{Second_Drafts/related_works}

\section{Methodology}

\input{Second_Drafts/methodology}

\section{Experiments}
\input{Second_Drafts/experiments}

\section{Conclusion}

\input{Second_Drafts/conclusion}

\vspace{-.3cm}
\ninept
\bibliographystyle{IEEEbib}
\bibliography{refs}

\end{document}

%% file: Second_Drafts/abstract.tex
In this paper, we propose a novel selection strategy for contrastive learning for medical images. On natural images, contrastive learning uses augmentations to select positive and negative pairs for the contrastive loss. However, in the medical domain, arbitrary augmentations have the potential to distort small localized regions that contain the biomarkers we are interested in detecting. A more intuitive approach is to select samples with similar disease severity characteristics, since these samples are more likely to have similar structures related to the progression of a disease.  To enable this, we introduce a method that generates disease severity labels for unlabeled OCT scans on the basis of gradient responses from an anomaly detection algorithm.  These labels are used to train a supervised contrastive learning setup to improve biomarker classification accuracy by as much as 6\% above self-supervised baselines for key indicators of Diabetic Retinopathy.

%% file: Second_Drafts/introduction.tex
\begin{figure}[t]
\centering
\includegraphics[width = \columnwidth]{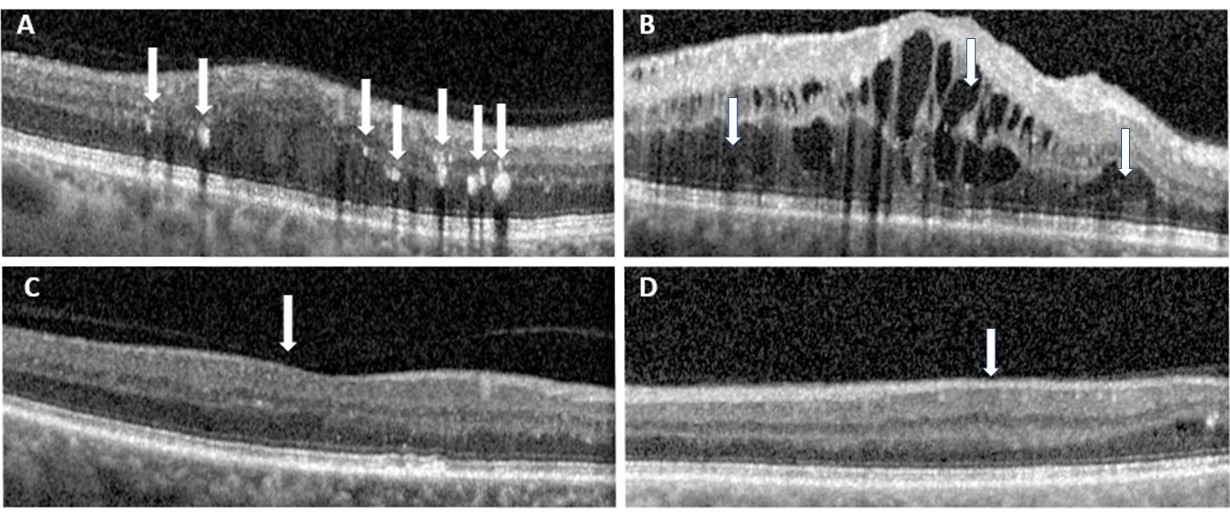}

\caption{OCT Scans of biomarkers from the Prime + TREX DME datasets. The biomarkers are A) Intraretinal Hyperreflective Foci (IRHRF), B) Intraretinal Fluid (IRF) and Diabetic Macular Edema (DME) C) Partially Attached Vitreous Face (PAVF), and  D) Fully Attached Vitreous Face (FAVF). The white arrows point out examples of where these biomarkers are found in the images. A discussion of biomarkers in OCT can be found at \cite{markan2020novel}.\vspace{-.3cm}}

\label{fig: examples}
\end{figure}

Diabetic Retinopathy (DR) is the leading cause of irreversible blindness among people aged 20 to 74 years old \cite{fong2004retinopathy}. In order to manage and treat DR, the detection and evaluation of biomarkers of the disease is a necessary step for any clinical practice \cite{markan2020novel}. Biomarkers refer to “any substance, structure, or process that can be measured in the body or its products and influence or predict the incidence of outcome or disease \cite{strimbu2010biomarkers}.” Biomarkers such as those in Figure \ref{fig: examples} are important indicators of DR and give ophthalmologists a fine-grained understanding of the manifestation of DR for individual patients.

Due to the importance of biomarkers in the clinical decision making process, much work has gone into deep learning methods to automate their detection directly from optical coherence tomography (OCT) scans \cite{kermany2018identifying}. A major bottleneck hindering this goal is the dependence of conventional deep learning architectures on having access to a large training pool. This dependency is not generalizable to the medical domain where biomarker labels are expensive to curate, due to the requirement of expert graders. In order to move beyond this limitation, contrastive learning  \cite{chen2020improved} has been one of the research directions to  leverage the larger set of unlabeled data to improve performance on the smaller set of labeled data. Contrastive learning approaches operate by creating a representation space through minimizing the distance between positive pairs of images and maximizing the distance between negative pairs. Traditional approaches like \cite{chen2020simple} generate positive pairs from taking augmentations from a single image and treating all other images in the batch as the negative pairs. This makes sense from a natural image perspective, but from a medical point of view the augmentations utilized in these strategies, such as gaussian blurring, can potentially distort small localized regions that contain the biomarkers of interest. Examples of regions that could potentially be occluded are indicated by white arrows in Figure \ref{fig: examples}. A more intuitive approach from a medical perspective would be to select positive pairs that are at a similar level of \emph{severity}. Images with similar disease severity levels share structural features in common that manifest itself during the progression of DR \cite{wang2021detection}. Hence, choosing positive pairs on the basis of severity can better bring together OCT scans with similar structural components in a contrastive loss. It is also possible to view more severe samples as existing on a separate manifold from the healthy trained images as shown in Figure \ref{fig: manifold}. From this manifold outlook of severity, some model response can be calculated as a severity score that indicates how far a sample is from the healthy manifold. To capture this intuition, we introduce the description of severity as ``the degree to which a sample appears anomalous relative to the distribution of healthy images." From this perspective, one way to measure severity is by formulating it as an anomaly detection problem where some response from a trained network can serve to identify the degree to which a sample differs from healthy images through a severity score.

We argue in this work that the appropriate model response to measure is the gradient from the update of a model. Gradients represent the model update required to incorporate new data. From this intuition, gradients have been shown to be able to represent the learned representation space from a model \cite{kwon2019distorted}, represent contrastive explanations between classes \cite{prabhushankar2020contrastive}, and perform contrastive reasoning \cite{prabhushankar2021contrastive}. Anomalous samples require a more drastic update to be represented than normal samples \cite{lee2020gradients}. Additionally, previous work \cite{agarwal2020estimating} showed that gradient information could be used to effectively rank samples into subsets that exhibit semantic similarities. Hence, in this work, we propose to use gradient measures from an anomaly detection methodology known as \texttt{GradCON} \cite{kwon2020backpropagated} to assign pseudo severity labels to a large set of unlabeled OCT scans. We utilize these severity labels to train an encoder with a supervised contrastive loss \cite{khosla2020supervised} and then fine-tune the representations learnt on a smaller set of available biomarker labels.  In this way, we leverage a larger set of easy to obtain healthy images and unlabeled data to improve performance in a biomarker classification task. The contributions of this paper are: \textbf{1)} We propose a framework to assign severity pseudo-labels to unlabeled OCT scans on the basis of gradient responses. \textbf{2)} We show the effectiveness of  these weak severity labels in a contrastive learning framework to improve biomarker classification performance.

%% file: Second_Drafts/related_works.tex
\begin{figure}[t]
\centering
\includegraphics[scale=.45]{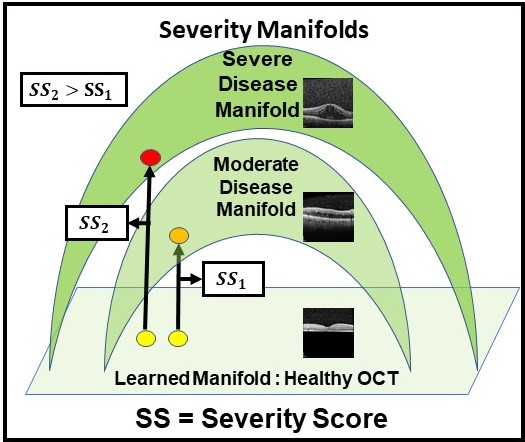}
\caption{From a healthy manifold learned from a trained auto-encoder, we can compute distance to the manifold of more severely diseased cases via a Severity Score (SS). Severity Score is calculated via some model response and increases as a sample is more anomalous compared to the learned healthy manifold.}

\label{fig: manifold}
\end{figure}

\textbf{Self-Supervised learning for OCT images:} Traditional deep learning approaches for OCT have relied on access to a large quantity of labeled data. For example, \cite{kermany2018identifying} showed how transfer learning could be applied to learn classification of retinal disease indicators. Similarly, \cite{temel2019relative} introduced a transfer learning scheme to detect pupillary defects.  Methods like this demonstrate the capability of deep learning in OCT, but their dependence on potentially expensive labels  makes them sub-optimal for the various constraints of the medical field. This has motivated self-supervised strategies such as \cite{rivail2019modeling} where a model was trained to learn the time interval between patient visits for characterizing disease progression in OCT. \cite{li2020self} also used patient information as well as modality-invariant features to detect Age-related macular degeneration (AMD). \cite{zhang2021twin} crated a dual framework leveraging labeled and unlabeled data for anomaly detection tasks in OCT scans. These works introduce self-supervised ideas to OCT, but differs from ours from the perspective of our pseudo-labeling process and the usage of a supervised contrastive loss. \\
\textbf{Contrastive Learning in Medical domain:} Previous work has shown that traditional contrastive learning approaches such as \cite{chen2020improved} and \cite{chen2020simple} work within the medical domain \cite{sowrirajan2020moco}. However, improvements have been sought by identifying more informed ways to choose positive instances. \cite{vu2021medaug} showed how choosing positive instances from the same patient can improve classification of Chest X-ray scans. Similarly, \cite{cheng2020subject} utilized a subject-aware contrastive learning with brain wave signals for the task of anomaly detection. \cite{zeng2021positional} used a supervised contrastive loss based on position within a volume. Each of these works are similar to ours in that they seek to introduce a medically consistent way to choose positive pairs from data for a contrastive loss. However, our work differs because it introduces a new methodology for choosing positive pairs of data based on weak disease severity labeling.

  \begin{figure*}
\centering
\begin{subfigure}{.5\textwidth}
  \centering
  \includegraphics[width=\columnwidth]{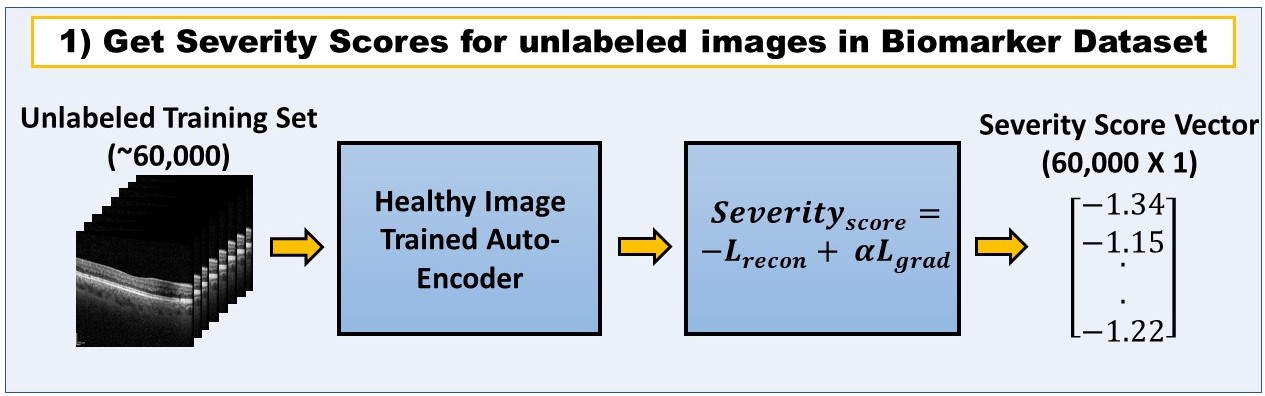}
  \caption{Compute the severity score for each unlabeled image.}
  \label{fig:sub1}
\end{subfigure}%
\begin{subfigure}{.5\textwidth}
  \centering
  \includegraphics[width=\columnwidth]{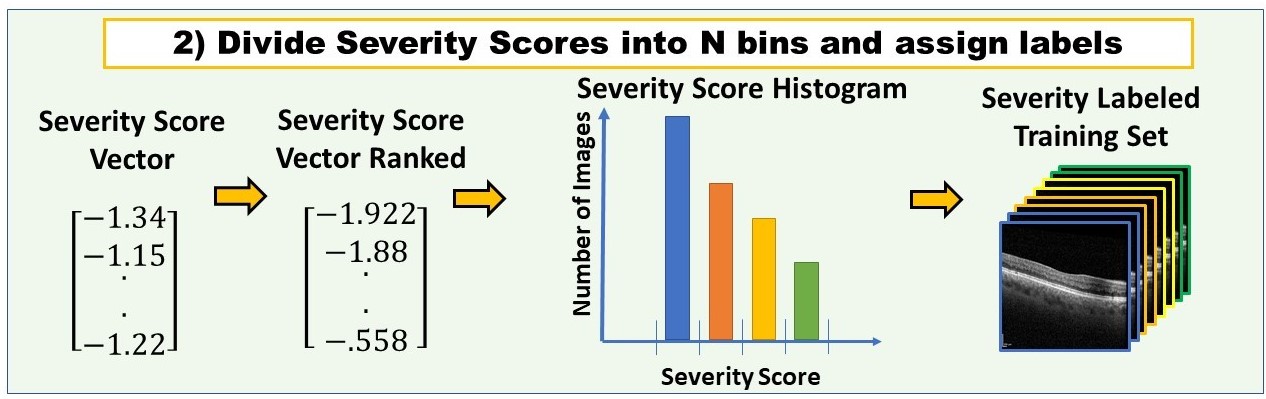}
  \caption{Divide severity score into $N$ bins and use to assign labels.}
  \label{fig:sub2}
\end{subfigure}
\caption{Overview of severity labeling methodology.}
\label{fig: severity}
\end{figure*}

\begin{figure}[ht]
\centering
\includegraphics[width=\columnwidth]{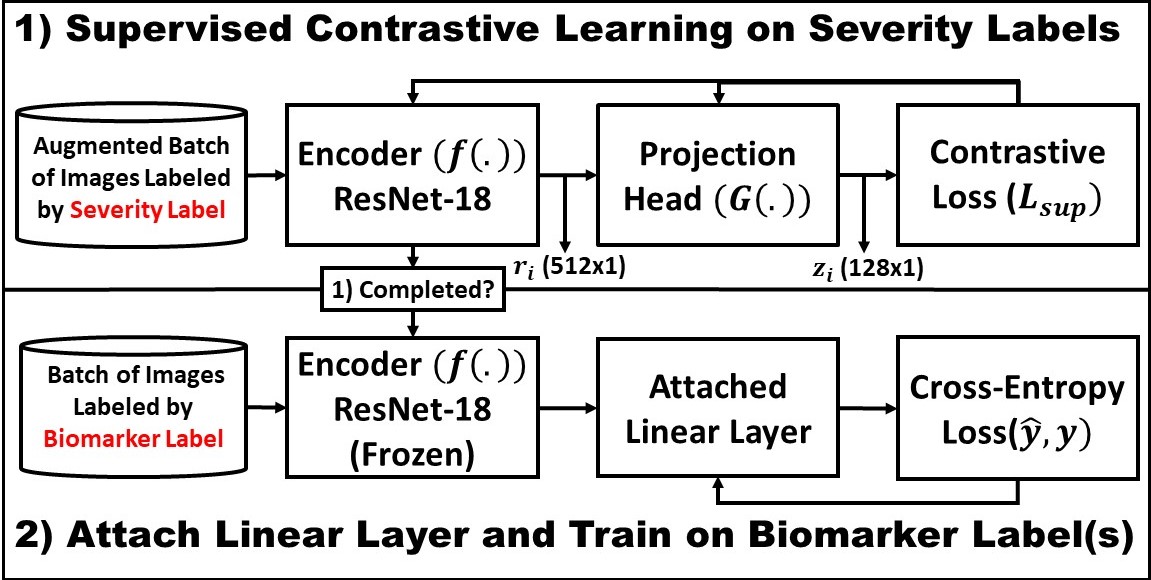}
\caption{ Overview of supervised contrastive learning and linear fine-tuning steps. 1) Supervised Contrastive Loss on generated severity labels from previously unlabeled data. 2) Attach linear layer and train on labeled biomarker data. \vspace{-.1cm}}
\label{fig: severity_supcon}
\end{figure}

%% file: Second_Drafts/methodology.tex
\subsection{Datasets}

The two datasets of interest are the healthy images from the Kermany dataset \cite{kermany2018large} and Prime + TREX DME \cite{payne2021long,hannah2021real}. We will refer to these datasets as the ``Healthy" dataset and the ``Biomarker" dataset respectively.
The  training set of the Biomarker dataset consists of approximately 60000 unlabeled OCT scans  and 7500 OCT scans with explicit biomarker labels from 76 different eyes. These OCT scans were collected as part of the \texttt{Prime} and \texttt{TREX DME} studies at the Retina Consultants of Texas (RCTX). For each labeled OCT scan in the Biomarker dataset,  a trained grader performed interpretation on OCT scans for the presence or absence of 20 different biomarkers. Among these biomarkers, 5 exist in balanced quantities to be used in training for the binary classification task of detecting the biomarker of interest. These biomarkers include those found in figure \ref{fig: examples}.
OCT scans from 76 eyes constitute the training set and the OCT scans from a remaining set of 20 eyes make up the test set. From these test OCT scans, random sampling was employed to develop an individualized test set for each of the 5 biomarkers used in our analysis. This resulted in a balanced test set, for each biomarker, where 500 OCT Scans have the biomarker present and 500 OCT Scans have the biomarker absent.

\subsection{Severity Label Generation}
The first step in our methodology is to learn the distribution of healthy OCT scans. To do this, the images from the Healthy dataset are resized to $224\times 224$ and used to train an auto-encoder through the \texttt{GradCON} methodology introduced by \cite{kwon2020backpropagated}. The intuition behind this work is to introduce a gradient constraint during training so that gradients from healthy images will align more closely together. This encourages images that deviate from the healthy distribution to have gradients that are more distinguishable. Once we have this trained auto-encoder, we follow the methodology detailed in Figure \ref{fig: severity}.  The first step is to get the severity score for all unlabeled images in the Biomarker dataset. To do this, we pass all unlabeled images to the input of the trained auto-encoder network and extract their corresponding severity score that is represented by equation \ref{eq:score}.
\begin{equation}
    \text{Severity Score (SS)} = -L_{recon} + \alpha L_{grad}
    \label{eq:score}
\end{equation}
 $L_{recon}$ is the mean squared error between an input $x$ and its reconstructed output $\hat{x}$. $L_{grad}$ is the average of the cosine similarity between the gradients of the target image and the reference gradients learned from training on the healthy dataset across all layers of the decoder. $\alpha$ was set to .03 for all experiments.  This results in every image having an associated severity score which constitutes a severity score vector. Once we have this  severity score vector, we move to step 2 in Figure \ref{fig: severity}, where the severity labels for each image are generated. This is done by ranking the severity scores in ascending order. These ranked scores are then divided into $N$ bins based on having a similar severity score to other images.  Images belonging to the same severity score bin are given the same severity label ($SL$) for use in the next stage of our setup. This results in the previously unlabeled data now having one of $N$ possible labels. Note that $N$ is a hyper-parameter that is explored experimentally in this paper. To get an intuitive sense of the quality of these severity labels we sample images randomly from the bins at the extreme ends of the histogram and end up with Figure \ref{fig: gradcon_dist}. It can be observed that the lower severity scores correspond with images that are healthier than those at the other end of the distribution. This makes sense as lower severity scores indicate that the image's gradients have a greater alignment with the gradients of the learned healthy distribution.

\subsection{Supervised Contrastive Loss on Severity Labels}

Once we have the severity labels ($SL$), we utilize the supervised contrastive loss \cite{khosla2020supervised} to bring embeddings of images with the same label together and push apart embeddings of images with differing labels. The overall setup is detailed through the flowchart shown in Figure \ref{fig: severity_supcon}. The first step described is how the training of the encoder with the supervised contrastive loss progresses.  Each image $x_{i}$ is passed through an encoder network $f(\cdot)$, that we set to be ResNet-18 ~\cite{he2016deep}, producing a $512\times 1$ dimensional vector $r_{i}$. This vector is further compressed through a projection head $G(.)$ which is set to be an multi-layer perceptron with a single hidden layer. This projection head is used to reduce the dimensionality of the representation and is discarded after training. The output of $G(.)$ is a $128\times 1$ dimensional embedding $z_{i}$. In this embedding space, the dot product of images with the same severity label (the positive samples) are maximized and those with different severity labels (the negative samples) are minimized. This takes the form of equation \ref{eq:sup} where positive instances for image $x_{i}$ come from the set $P(i)$ and positive and negative instances come from the set $A(i)$. $\tau$ is a temperature scaling parameter set to .07 for all experiments. $z_{p}$ refers to the embeddings of the set of positive instances and $z_{a}$ indicates embeddings of all positive and negative instances.
\begin{equation}
    \label{eq:sup}
     L_{sup} = \sum_{i\in{I}} \frac{-1}{|P(i)|}\sum_{p\in{P(i)}}log\frac{exp(z_{i}\cdot z_{p}/\tau)}{\sum_{a\in{A(i)}}exp(z_{i}\cdot z_{a}/\tau)}
\end{equation}

After training the encoder, we move to step 2 of the methodology described in Figure \ref{fig: severity_supcon} where the model explicitly learns to detect biomarkers. To do this, the weights of the encoder are frozen and a linear layer is appended to the output of the encoder. We utilize the subset of the Biomarker dataset with biomarker labels for fine-tuning on top of the representation space learnt in step 1 of Figure \ref{fig: severity_supcon}. A biomarker is chosen to be the one we are interested in and a linear layer is trained using cross-entropy loss between a predicted output $\hat{y}$ and a ground truth label $y$ to learn to detect the presence or absence of the biomarker in the image.

%% file: Second_Drafts/experiments.tex
\begin{table*}[h]
\centering
\begin{tabular}{@{}ccccccc@{}}

\toprule
\multicolumn{7}{c}{Severity Label Training Results (Accuracy / F1-Score)}   
\\ \midrule
\multicolumn{1}{|c|}{Method} & \multicolumn{1}{c|}{IRF} & \multicolumn{1}{c|}{DME} & \multicolumn{1}{c|}{IRHRF} & \multicolumn{1}{c|}{FAVF} & \multicolumn{1}{c|}{PAVF} & \multicolumn{1}{c|}{Multi-Label} \\ \midrule
SimCLR \cite{chen2020simple}                    & 75.13\% / .715  & 80.61\% / .772 & 59.03\% /.675 & 75.43\% / .761 & 52.69\% / .249 & .754  \\
PCL \cite{li2020prototypical}                    & \textbf{76.50}\% / .717  & 80.11\% / .761 & 59.1\% /.683 & \textbf{76.30}\% / .773 & 51.40\% / .165 & .767  \\
Moco v2 \cite{chen2020improved}                   & 76.00\% / .720      &  82.24\% / .793       & 59.6\% / .692       &  75.00\% / \textbf{.784}      &  52.5\% / .201   & .769  \\ \bottomrule
$SL_{5000}$ & 75.20\% / .698  &81.46\% / .786 & \textbf{66.83}\% / .695 & 75.39\% / .756 & 54.7\% / .314 & \textbf{.774} \\
$SL_{10000}$ & 75.46\% / .732 & 82.35\% / .796 & 62.00\% / .700 & 73.56\% / .722 & \textbf{56.69}\% / \textbf{.370} & .771 \\
$SL_{15000}$ & 74.33\% / .701 & \textbf{84.52}\% / \textbf{.831} & 59.86\% / .691 & 72.03\% / .726 & 55.56\% / .322 & .765  \\
$SL_{20000}$ & 76.13\% / \textbf{.733} & 81.12\% / .788 & 62.93\% / \textbf{.703} & 71.90\% / .745 & 54.03\% / .258 & .766 \\
 \bottomrule
\end{tabular}
\caption{Performance of severity labeling and supervised contrastive learning approach on individual biomarkers. Multi-Label is the average AUC from the multi-label classification task. $SL_{N}$ refers to training an encoder after having divided the training set into $N$ severity label bins. }
\label{tab:main_table}
\end{table*}

The proposed strategy is compared against three state of the art self-supervised strategies that make use of the unlabeled data. The architecture was kept constant as ResNet-18 across all experiments. Augmentations included random resize crop to a size of 224, horizontal flip, color jitter, and normalization to the mean and standard deviation of the respective dataset. The batch size was set at 64. Training was performed for 25 epochs in every setting. A stochastic gradient descent optimizer was used with a learning rate of 1e-3 and momentum of .9. 
Accuracy and F1-score was recorded for the testing  of performance on each individual biomarker. Additionally, we assess the method's capability across all biomarkers by utilizing a mean AUC metric within a multi-label classification setting for the labels of all 5 biomarkers at the same time.
\begin{table}[]
\centering
\begin{tabular}{@{}cc@{}}

\toprule
\multicolumn{2}{c}{Anomaly Detector Ablation Study}                       \\ \midrule
\multicolumn{1}{|c|}{Method} & \multicolumn{1}{c|}{Multi-Label} \\ \midrule

MSP \cite{hendrycks2016baseline} & .764 \\
ODIN \cite{liang2017enhancing} & .761 \\
Mahalabonis \cite{lee2018simple} & .772 \\ \bottomrule
SL & \textbf{.774} \\

 \bottomrule
\end{tabular}
\caption{Analysis of using different anomaly detectors to generate severity labels. For each method scores were generated for each unlabeled image in the Biomarker dataset and discretized into 5000 bins. \vspace{-.3cm}}
\label{tab:anomaly}
\end{table}

Overall, choosing of the severity bin hyperparameter $N$ has significant influence on performance improvements on both multi-label classification performance as well as detection performance on individual biomarkers. Table \ref{tab:main_table} represents the complete results for this study. $SL_{N}$ represents the encoder trained after dividing the unlabeled pool into $N$ severity label bins.  This is repeated for different number of severity bin divisions ranging from 5000 to 20000.  A larger number of bins means the images within each bin are more likely to share structures in common, but at the trade-off of having fewer positive instances during training. A more moderate choice for number of severity bins, such as 5000 or 10000, led to improved performance in multi-class classification compared to all baselines. However, it was also observed that choosing larger or smaller values for the number of bins resulted in best performance for specific biomarkers. For example, the best performing values for DME and IRF resulted at the higher number of severity bins of 15000 and 20000. Additionally, the best result for PAVF was at a severity bin value of 10000. The variation in performance for different bin numbers can also be understood from the perspective of how fine-grained individual biomarkers are. IRF and DME manifest themselves more distinctively than IRHRF, FAVF, or PAVF. Therefore, it's possible that a lower number of more closely related positives may be better for identifying these distinctive features. However, in the case of more difficult to identify biomarkers, it may be the case that some level of diversity in the positives is necessary to identify them from the surrounding structures in the OCT scan.

We also studied the effect of using other anomaly detection methods to generate severity scores. We trained a classifier using the labeled data from the Biomarker dataset. Using the output logits from this classifier, we can generate anomaly scores for each the methods shown in Table \ref{tab:anomaly}. These scores are then processed in the same way as Figure \ref{fig: severity} to generate labels for the unlabeled dataset. These labels are used in the same way as the methodology introduced in this paper to train an encoder with a supervised contrastive loss. Performance is measured by the average AUC value from a multi-label classification setting. We observe that our method with the same level of discretization out-performs all other anomaly detection methodologies.

%% file: Second_Drafts/conclusion.tex
\begin{figure}[]
\centering
\includegraphics[width = \columnwidth]{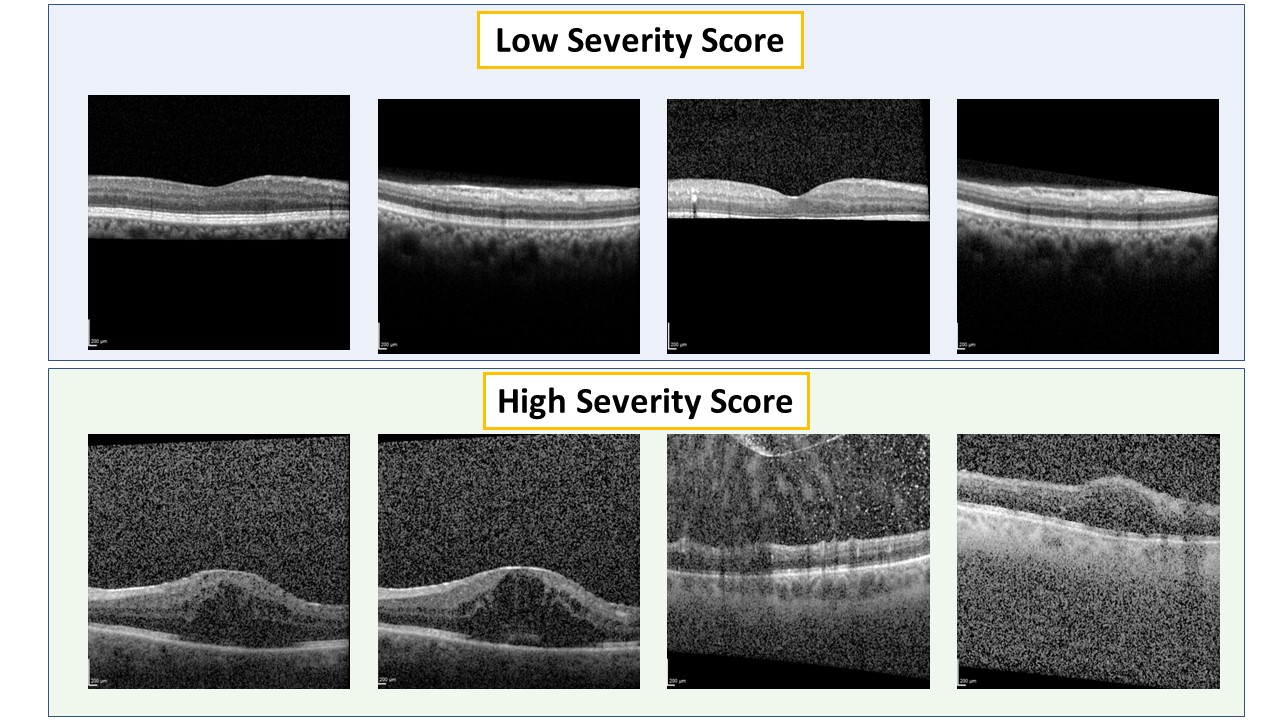}

\caption{ Visual representation of OCT scans with high and low severity scores.\vspace{-.3cm}}
\label{fig: gradcon_dist}
\end{figure}

Contrastive learning is useful to the medical domain because it allows us to integrate a large pool of unlabeled data into the training process. It relies on methods of choosing good positive and negative pairs of images. We show in this paper that one such approach is to choose pairs based on how anomalous they appear relative to a learned healthy class distribution. We show that defining this degree of abnormality through a gradient response leads to semantically interpretable clusters under a common label that can be used within a supervised contrastive learning framework. Training on these labels leads to an improved representation space that out-performs state of the art self-supervised methods.